\documentclass[letterpaper, 10 pt, conference]{ieeeconf}  

\IEEEoverridecommandlockouts                              

\usepackage{graphicx}
\usepackage{xcolor}
\usepackage{fancyhdr}

\title{\LARGE \bf
Deeply Supervised Active Learning for Finger Bones Segmentation 
}

\author{Ziyuan Zhao$^{1}$, Xiaoyan Yang$^{2}$, Bharadwaj Veeravalli$^{2}$, and Zeng Zeng$^{1,3}$
\thanks{$^{1}$Ziyuan Zhao and Zeng Zeng are with  Institute for Infocomm Research (I2R), Agency for Science, Technology and Research (A*STAR), Singapore 
}
\thanks{$^{2}$Xiaoyan Yang and Bharadwaj Veeravalli are with National University of Singapore, Singapore}
\thanks{$^3$ Corresponding author. The work was supported by Singapore-China NRF-NSFC Grant (Grant No. NRF2016NRF-NSFC001-111)}
}

\begin{document}

\maketitle
\thispagestyle{empty}
\pagestyle{empty}


\thispagestyle{fancy}
\fancyhead{}
\lfoot{}
\lfoot{\scriptsize{Copyright 2020 IEEE. Published in the 2020 42nd Annual International Conference of the IEEE Engineering in Medicine and Biology Society (EMBC), scheduled for July 20-24, 2020 at the Montréal, Canada. Personal use of this material is permitted. However, permission to reprint/republish this material for advertising or promotional purposes or for creating new collective works for resale or redistribution to servers or lists, or to reuse any copyrighted component of this work in other works, must be obtained from the IEEE. Contact: Manager, Copyrights and Permissions / IEEE Service Center / 445 Hoes Lane / P.O. Box 1331 / Piscataway, NJ 08855-1331, USA. Telephone: + Intl. 908-562-3966.}}
\rfoot{}

\begin{abstract}
Segmentation is a prerequisite yet challenging task for medical image analysis. In this paper, we introduce a novel deeply supervised active learning approach for finger bones segmentation. The proposed architecture is fine-tuned in an iterative and incremental learning manner. In each step, the deep supervision mechanism guides the learning process of hidden layers and selects samples to be labeled. Extensive experiments demonstrated that our method achieves competitive segmentation results using less labeled samples as compared with full annotation.
\newline

\indent \textit{Clinical relevance}— The proposed method only needs a few annotated samples on the finger bones task to achieve comparable results in comparison with full annotation, which can be used to segment finger bones for medical practices, and generalized into other clinical applications.
\end{abstract}

\section{INTRODUCTION}
Image segmentation plays an important role in multiple applications of medical image analysis, {\textit{e}.\textit{g}.}, medical diagnosis, surgical planning and pathological analysis~\cite{8803074, larson2017performance}. Recently, deep learning based methods have achieved great success in image segmentation. However, different from segmenting natural scene images, medical image segmentation, such as hand bones segmentation, is challenging, because it is difficult to obtain medical image datasets with full annotation. On one hand, most biomedical image annotations require extensive domain knowledge and skills, which means only well trained experts can label the data. On another hand, manual labeling is tedious, costly and time-consuming. These impeded the development of deep learning in medical imaging.

To alleviate the aforementioned challenges, some self-supervised~\cite{SeseNet} or semi-supervised segmentation algorithms~\cite{NIPS2015_5858, 8834460} have been proposed, which leverage the model-generated labels (pseudo labels) to boost the performance of deep learning models iteratively. However, there is uncertainty about how to evaluate the quality of pseudo labels. Besides, in practice using a small number of accurate labels is more meaningful than using many noisy and uncertain pseudo labels. This motivates us to seek a solution to the problem of selecting data samples to be labeled for high-quality performance.

Accurate selection of informative and valuable samples to label involves the problem of Active Learning~(AL). AL has been well researched on various vision tasks~\cite{7780682, wang2016cost, kading2015active, hasan2015context}, but only a few studies have been reported on medical imaging. The method AIFT (active, incremental fine-tuning) proposed in~\cite{zhou2017fine} integrates active learning and transfer learning (fine-tuning) into a single framework, which selects informative samples based on entropy and diversity. The deep active learning framework proposed in~\cite{yang_suggest} first trains a set of Fully Convolutional Networks~(FCNs) with little training data and then seeks informative samples based on the uncertainty among the trained FCNs and similarity between images. However, these methods based on uncertainty and similarity cannot be well embedded with final metrics. This further motivates us to study the following question: \emph{Is it possible to improve the finger bones segmentation performance by guiding deep active learning directly?}

In this work, we present a new framework that combines deep learning and active learning to reduce the annotation effort and improve the model performance by injecting deep supervision into hidden layers of the segmentation model. With a small number of training data, the U-Net is trained under deep supervision and then boosted iteratively. At the end of each stage, we exploit the deep supervision mechanism to extract useful information to decide which samples to label for the next batch. The U-Net is continuously fine-tuned by incrementally enlarging the training dataset with newly annotated samples. Experimental results show that the proposed method achieves comparable performance with less labeling cost as opposed to full annotation. 

\begin{figure*}[htb]
    \centering
    \includegraphics[width=16cm]{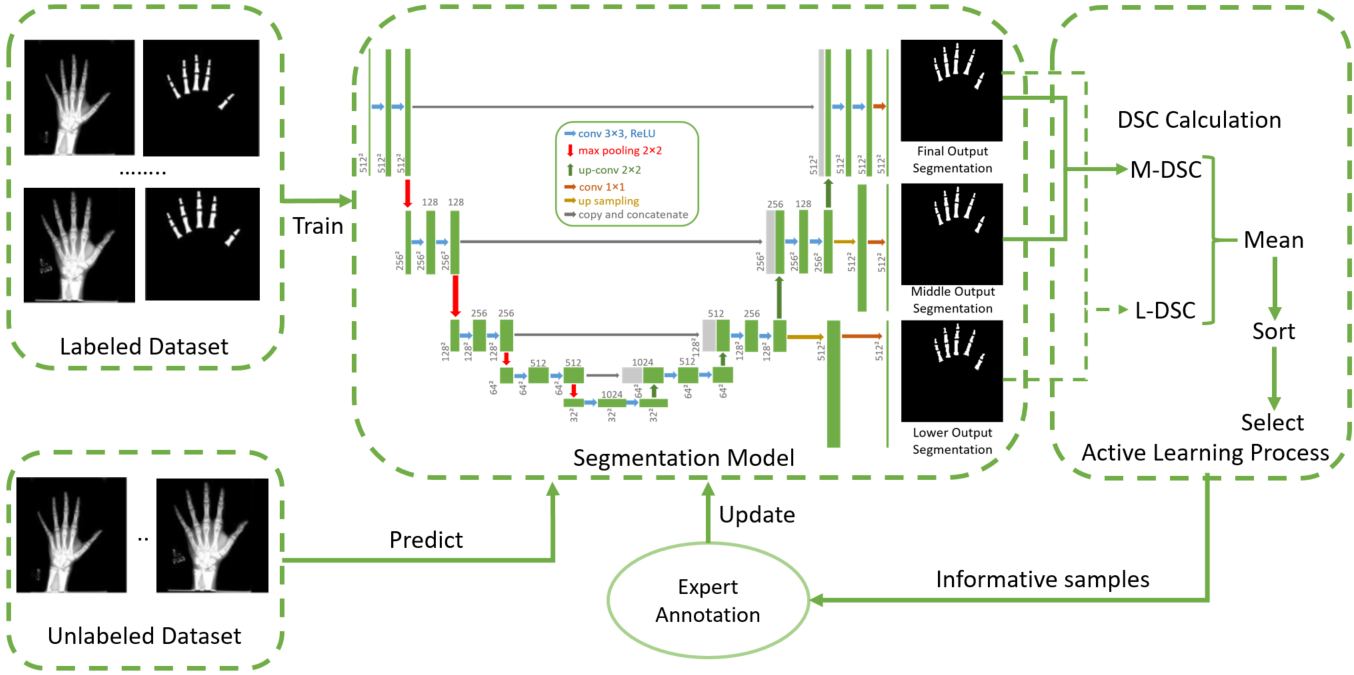}
    \caption{Overall framework of our proposed method: the segmentation model is first trained using the labeled dataset, and the unlabeled dataset is applied on the trained networks to get multi-level segmentation outputs as the inputs of active learning process. In the active learning process, the dice coefficient between different segmentation outputs are calculated and averaged as the evaluation criteria for informative samples to label in next iteration. Informative samples are selected for annotation and the updated labeled dataset is used for the training process of the next batch.}
    \label{fig:pipeline}
\end{figure*}
\section{RELATED WORK}

Deep learning has dominated computer vision over the past few years, which now is a primary option for image segmentation, especially for medical image segmentation. So far, many work~\cite{7803544,long2015fully, Unet} in the literature adopt an ``encoder-decoder” structure, in which, the resolution of the input is compressed and then, recovered by the encoder and the decoder sequentially. Among them, U-Net is commonly used in medical image segmentation, which concatenates multi-scale feature maps in the encoding stage with upsampled feature maps in the decoding stage. Such a design can be trained using very few annotated images and achieves outstanding performance in medical image segmentation.

However, different from other tasks, such as classification and regression, pixel-wise annotation is required for deep learning in medical image segmentation. Generally, only trained biomedical experts can annotate the data, which also needs extensive manual efforts. To address these issues, some deep learning frameworks~\cite{NIPS2015_5858, 8834460} based on semi-supervised learning are developed for pseudo-labeling and finer segmentation. But these methods cannot well define how to select informative images for annotation. Therefore, several work based on active learning~\cite{zhou2017fine, yang_suggest} were proposed to seek the answer to the problem. Zhou~\emph{et al.}~\cite{zhou2017fine} proposed an active learning method for medical image segmentation called ``AIFT”, in which, active learning and transfer learning are integrated to boost the entire framework for sample selection. Yang~\emph{et al.}~\cite{yang_suggest} trained a set of FCNs to estimate the uncertainty among them as annotation criterion. These methods are based on uncertainty and similarity, which cannot be directly linked to the final evaluation. It is well noted that using some regularization techniques in the learning process, the information in hidden layers can be used to improve the model performance. For example, the deep supervision mechanism in~\cite{DOU201740} was introduced for liver segmentation, in which, the dense predictions are obtained from the hidden feature maps of some hidden layers for calculating classification errors, which can simultaneously speed up the optimization process and improve the discrimination capability. We speculate that the feature maps of hidden layers may provide guidance for active learning because of the spatial similarity between different feature maps. Therefore, we inject deep supervision into some hidden layers of U-Net and design a novel active learning strategy for finger bones segmentation.

\section{METHODOLOGY}

Our active learning architecture is illustrated in Fig.~\ref{fig:pipeline}, which can be summarized as follows. U-Net with deep supervision (deeply supervised U-Net) is adopted as the segmentation model, where additional multi-level deep supervision is injected into some hidden layers to improve the discrimination ability of the model. The model is trained on a small amount of labeled data firstly. In each iteration of training, the predicted masks of the remaining unlabeled data from hidden layers and final layers can be outputted by segmentation, which serves as the input of the active learning model. In the active learning model, the quality of the samples can be estimated by the consistency between the results of hidden layers and the final layer, and the indices of the informative samples are exported for expert annotation. To provide the ground truth of the informative samples, the process of expert annotation is simulated using the fully annotated dataset, then the training set is updated and the segmentation model is fine-tuned with the updated dataset iteratively and incrementally.

\subsection{Deeply Supervised U-Net}
The segmentation model is shown in Fig.~\ref{fig:pipeline}. We follow the basic architecture of U-Net, The deep supervision mechanism is involved via the twentieth and twenty-third layers, namely lower layer and middle layer, in which, we upscale some low-level and middle-level feature maps of lower layer and middle layer respectively on the decoding stage using additional upsampling layers, and the softmax layer is applied on these layers as the final layer to obtain predicted masks for final loss calculation. Let $W$ be the weights of the U-Net, and $w^{l}$, $w^{m}$, $w^{f}$ be the weights of three classifiers of lower layer, middle layer and final layer, respectively, then the cross-entropy loss function of a layer is defined as

\begin{equation}
\mathcal{L}(\mathcal{X} ; W)=\sum_{x_{i} \in \mathcal{X}}-\log p\left(t_{i} | x_{i} ; W\right),
\end{equation}
where $\chi$ represents the training samples and $p\left(y_{i}=t\left(x_{i}\right) | x_{i} ; W, w^{c}\right)$ is the probability of target class label $t\left(x_{i}\right)$ corresponding to sample $x_{i} \in \chi$, in which, $c\in\{l, m, f\}$ represents the index of the classifiers;  

Finally, the total loss function can be defined as:
\begin{equation}
\mathcal{L}\left(\chi ; W, w^{l}, w^{m}, w^{f}\right)=\sum_{c \in\{l, m, f\}} \alpha_{c} \mathcal{L}_{c}\left(\chi ; W, w^{c}\right),
\end{equation}
where $\alpha_{l}$, $\alpha_{m}$, $\alpha_{f}$ are the weights of the associated classifiers. To control the strength of different terms, in our experiments, $\alpha_{l}$, $\alpha_{m}$ and $\alpha_{f}$ were specified as 0.1, 0.3 and 0.6, empirically.

In this way, different classifiers are simultaneously performed, which enforce the hidden layers to learn discriminative features without introducing many parameters. 

\begin{figure}[htb]
    \centering
    \includegraphics[width=8.1cm]{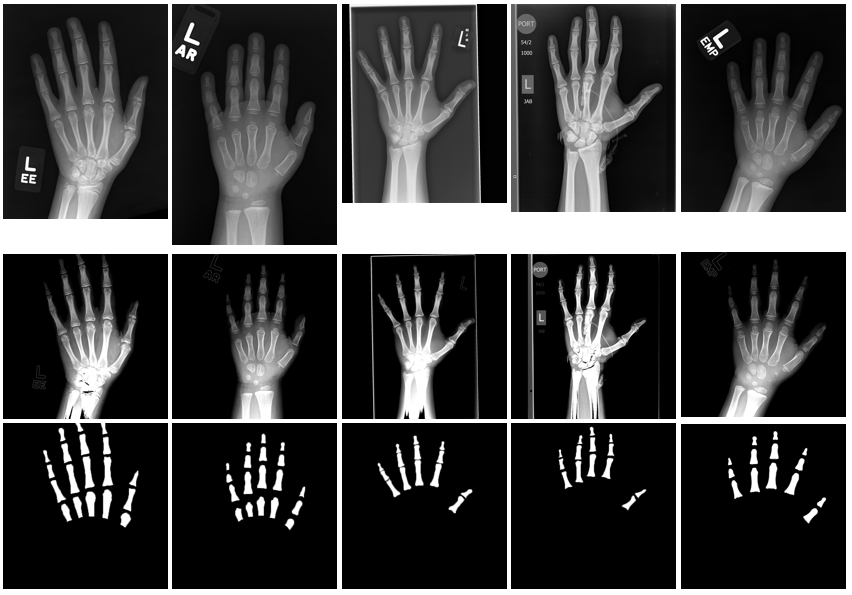}
    \caption{Example of radiographs in the dataset in different size, contrast and brightness: original images (first row), preprocessed images (second row) and binary finger bone masks  (third row).}
    \label{fig:example}
\end{figure}


\subsection{Active Learning Model}

The core of active learning is to define criteria to select informative samples. We propose a new measure of consistency in segmentation tasks for sample selection. Deep supervision in hidden layers can improve the discrimination ability of the model, and more accurate prediction from hidden layers also means better learned features, which may be similar to the output of the last layer. Moreover, different layers of a good model can perform well and consistently on high-informative samples. Therefore, the segmentation results of the hidden layers (lower layer and middle layer) are extracted to calculate the Dice's Coefficient (DSC) with the results of the final layer, called L-DSC and M-DSC, respectively. The DSC is calculated as

\begin{equation}
\label{eq:DSC}
D S C=\frac{2|X \cap Y|}{|X|+|Y|},
\end{equation}
where $X$ and $Y$ represent the cardinalities of the two sets. 

Finally, the mean of L-DSC and M-DSC serves as a proxy of the quality for each sample, which will be larger on informative samples. In another word, the images with larger mean of L-DSC and M-DSC will be selected to update the training process.

\section{EXPERIMENTS}
\subsection{Dataset and Evaluation Protocol}
The dataset is from the 2017 RSNA Bone Age Prediction Challenge~\cite{larson2017performance}, which includes $12611$ hand radiographs ($6833$ for male and $5778$ for female). Some samples are shown in Fig.~\ref{fig:example}. To clean the data, we first applied intensity normalization to improve the brightness and sharpness of bones. Then the images were cropped and resized to $600 \times 600$ pixels. Considering the size of the dataset and manual costs, a small percentage of the dataset is sampled in our experiments to validate the effectiveness of the proposed method. For a better generalization of the segmentation, the dataset was grouped in 19 year-based age groups (0-1, 1-2, 2-3, \ldots ,18-19), then 11 cases were randomly selected from each year group to form the small subset of the training set ($11\times19 = 209$) for experiments. Finally, the dataset is split into a training (139) /validation (20) /testing (50) set randomly.

To validate the effectiveness of the proposed pipeline, experiments were done in the 209 images with ground truth masks. We first trained a deeply supervised U-Net using 10 images randomly selected from the training set. After each step, according to the active learning model, the top 10 informative images with annotations are put into deeply supervised U-Net to update the training process. The process is called ``DS-U-Net query”. Random query (randomly selecting 10 images to label) is also implemented using the same dataset for comparison.



\begin{figure}[htb]
    \centering
    \includegraphics[width = 8.3cm]{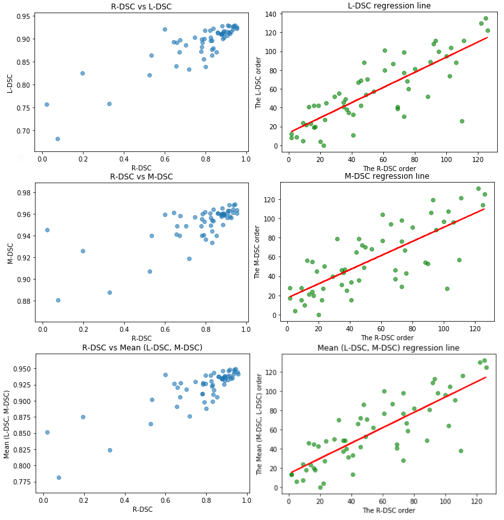}
    \caption{The relationships between R-DSC and L-DSC (M-DSC, mean of them). On left side, let x be the R-DSC, y be the L-DSC (M-DSC, mean of them). On right side, DSCs are sorted by descending order, let x be the order of R-DSC, y be the order of e L-DSC (M-DSC, mean
of them). We draw the scatter plots based on (x, y). Linear regression was carried out with the scatters on the right side. The regression coefficient is 0.80, 0.73 and 0.80 from top to bottom.}
    \label{fig:visual}
\end{figure}

\subsection{Results and Discussions}

To explore the relationships between real DSC (R-DSC) and L-DSC / M-DSC / mean of them, in the iterative training process, we randomly select 10 samples from each step and put them into the model for prediction. From Fig.~\ref{fig:visual}, we can see that the L-DSC / M-DSC / Mean of them are positively correlated with the real DSC. Because there are slight changes and fluctuations of DSC over the training process, we sort the DSC by descending order in each iteration, and the scatter plots and trend lines between the order of L-DSC / M-DSC / mean of them and the order of the real DSC are drawn. There are strong relationships between these orders, which proves the feasibility of the proposed criteria in active learning. Therefore, the segmentation accuracy can be correctly estimated by the active learning model, and the most informative slices can be selected.

\begin{figure}[htb]
    \centering
    \includegraphics[width = 8.5cm]{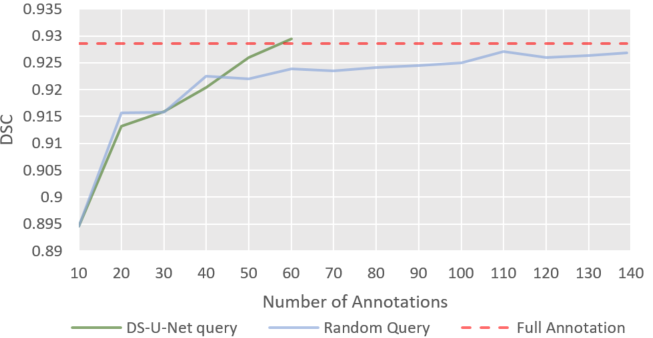}
    \caption{Results of experiments. This figure shows that our approach can get a comparable results only using 60 annotations (43.16\% of the whole dataset).}
    \label{fig:result}
\end{figure}

The experimental results on the testing data are shown in Fig.~\ref{fig:result}. We can see that out strategy is better than random query strategy, only using 43.16\% to achieve comparable results as compared with full annotation, which benefits from the deep supervision on hidden layers. This proves that the deep supervision mechanism not only helps to improve the discrimination ability of the segmentation model, but also serve as a criterion for active learning.


\section{CONCLUSIONS AND FUTURE WORK}
In this paper, we propose a novel active learning strategy for finger bones segmentation. Auxiliary deep supervision is injected after certain hidden layers of a deep U-Net model, which not only help convergence, but also serves as a criterion for active learning. The deeply supervised U-Net is first initialized on a subset of training set. Then it is updated and fine-tuned iteratively and incrementally. In each iteration, informative samples are selected based on the consistency between final masks and masks from hidden layers. Experimental results show that the proposed method outperforms random query and obtains comparable results only using 43.16\% samples as opposed to full annotation. Moreover, observations on the relationships between final masks and masks from hidden layers proves the effectiveness of the proposed active learning.

In the future research, we can implement it for semi-supervised learning and pseudo labeling. Deep supervision can provide guidance to the training process instead of introducing many parameters, and potentially reducing the cost of model design and implementation.

\bibliographystyle{IEEEbib}
\bibliography{refs.bib}

\end{document}